\documentclass[conference]{IEEEtran}

\usepackage{booktabs} % For formal tables
\usepackage{tabularx}
\usepackage{xspace}
\usepackage{enumitem}
\usepackage{xcolor,colortbl}
\usepackage{amsmath}
\usepackage{amssymb}
\usepackage{multirow}
\usepackage{hyperref}
\usepackage{graphicx}
\usepackage[center]{subfigure}
\usepackage[ruled,linesnumbered]{algorithm2e} % algorithm
\usepackage[]{algorithmic} % algorithm\textbf{}
\usepackage{tablefootnote}
\usepackage{threeparttable}
\usepackage{makecell}
\usepackage[font={small}]{caption}
\usepackage{diagbox}
\usepackage{empheq}
\usepackage{algorithmic}
\usepackage{ulem}

\begin{document}
	
\title{Autonomous Graph Mining Algorithm Search\\with Best Speed/Accuracy Trade-off}

\author{\IEEEauthorblockN{Minji Yoon}
	\IEEEauthorblockA{Carnegie Mellon University\\
		minjiy@cs.cmu.edu}
	\and
	\IEEEauthorblockN{Th\'{e}ophile Gervet}
	\IEEEauthorblockA{Carnegie Mellon University\\
		tgervet@andrew.cmu.edu}
	\and
	\IEEEauthorblockN{Bryan Hooi}
	\IEEEauthorblockA{National University of Singapore\\
		bhooi@comp.nus.edu.sg}
	\and
	\IEEEauthorblockN{Christos Faloutsos}
	\IEEEauthorblockA{Carnegie Mellon University\\
		christos@cs.cmu.edu}
}

\maketitle
\newtheorem{lemm}{Lemma}
\newtheorem{defi}{Definition}
\newtheorem{theo}{Theorem}
\newtheorem{assumption}{Assumption}
\newtheorem{observation}{Observation}
\newtheorem{problem}{Problem}
\newtheorem{algo}{Algorithm}
\newtheorem{property}{Property}

\renewcommand{\frame}{\textsc{UnifiedGM}\xspace}
\newcommand{\method}{\textsc{AutoGM}\xspace}

\newcommand{\mat}[1]{\mathbf{#1}}
\newcommand{\set}[1]{\mathbf{#1}}
\newcommand{\vect}[1]{\mathbf{#1}}

\newcommand{\A}{A} 
\newcommand{\NA}{\tilde{A}} 
\newcommand{\NAT}{\tilde{A}^{\top}} 
\newcommand{\D}{D} 
\newcommand{\ND}{\tilde{D}}
\newcommand{\I}{I} 

\newcommand{\X}{\mathcal{X}}
\newcommand{\Data}{\mathcal{D}} 
\newcommand{\Z}{\mathbb{Z}} 
\newcommand{\R}{\mathbb{R}} 

\newcommand{\f}{f_{GM}}

\newcommand{\note}[1]{\textcolor{red}{#1}}
\newcommand{\bh}[1]{\textcolor{blue}{#1}}

\newcommand{\codeurl}{\footnote{\url{https://github.com/minjiyoon/ICDM20-AutoGM}}}

\begin{abstract}
Graph data is ubiquitous in academia and industry, from social networks to bioinformatics.
The pervasiveness of graphs today has raised the demand for algorithms that can answer various questions: Which products would a user like to purchase given her order list? Which users are buying fake followers to increase their public reputation?
Myriads of new graph mining algorithms are proposed every year to answer such questions --- each with a distinct problem formulation, computational time, and memory footprint.
This lack of unity makes it difficult for a practitioner to compare different algorithms and pick the most suitable one for a specific application.
These challenges --- even more severe for non-experts ---  create a gap in which state-of-the-art techniques developed in academic settings fail to be optimally deployed in real-world applications.

To bridge this gap, we propose \method, an automated system for graph mining algorithm development.
We first define a unified framework \frame that integrates various message-passing based graph algorithms, ranging from conventional algorithms like PageRank to graph neural networks.
Then \frame defines a search space in which five parameters are required to determine a graph algorithm.
Under this search space, \method explicitly optimizes for the optimal parameter set of \frame using Bayesian Optimization. 
\method defines a novel budget-aware objective function for the optimization to incorporate a practical issue --- finding the best speed-accuracy trade-off under a computation budget --- into the graph algorithm generation problem.
Experiments on real-world benchmark datasets demonstrate that \method generates novel graph mining algorithms with the best speed/accuracy trade-off compared to existing models with heuristic parameters.

\end{abstract}

\begin{IEEEkeywords}
	automation, unified framework, optimization
\end{IEEEkeywords}

\section{Introduction}
\label{sec:introduction}
Many real-world problems are naturally modeled using graphs: who-buys-which-products in online marketplaces~\cite{yoon2019fast},
who-follows-whom in social networks~\cite{page1999pagerank,yoon2018tpa}, and protein relationships in biological networks~\cite{brohee2006evaluation, vazquez2003global}. 
Graph mining provides solutions to practical problems such as classification of web documents~\cite{yao2019graph, zamir1998web}, clustering in market segmentation~\cite{strehl2000scalable}, recommendation in streaming services~\cite{bennett2007netflix}, and fraud detection in banking~\cite{drezewski2015application, michalak2011graph}.

A dizzying array of new graph mining algorithms is introduced every year to solve these real-world problems, giving rise to the question: Which algorithm should we choose for a specific application?
Graph mining algorithms designed to solve the same task often have distinct conceptual formulations.
Concretely, to estimate the similarity between two nodes --- in social recommender systems for example --- classical graph mining algorithms (like Personalized PageRank~\cite{bahmani2010fast}) compute similarity scores by iterating a closed-form expression, while graph neural network algorithms ~\cite{wu2019comprehensive} first learn node embeddings using deep learning, then estimate similarity scores with a distance metric in this embedding space. 
This lack of unity makes it hard for practitioners to determine which aspect of a method contributes to differences in computation time, accuracy, and memory footprint --- significantly complicating the choice of the algorithm.
Currently, selecting a graph mining algorithm suitable for a specific task among dozens of candidates is a resource-intensive process requiring expert experience and brute-force search.

To mitigate the cost and complexity of the algorithm selection process, the machine learning community has developed AutoML~\cite{kandasamy2018neural,liu2018darts} --- which automates the process of algorithm selection and hyperparameter optimization.
%The AutoML paradigm has recently seen some meaningful successes in the search for neural network architectures~\cite{kandasamy2018neural,liu2018darts}.
The success of AutoML depends on the size of the search space: it should be small enough to be tractable in a reasonable amount of time.
However, AutoML techniques cannot be directly applied to graph mining because
the hyperparameter search space is not even defined due to the lack of unity among graph mining algorithms.

\begin{figure*}[htbp]
	\centering
	\vspace{-5mm}
	\subfigure[Accuracy constraints on the Citeseer dataset]
	{
		\label{fig:citeseer:acc}
		\includegraphics[width=.48\linewidth]{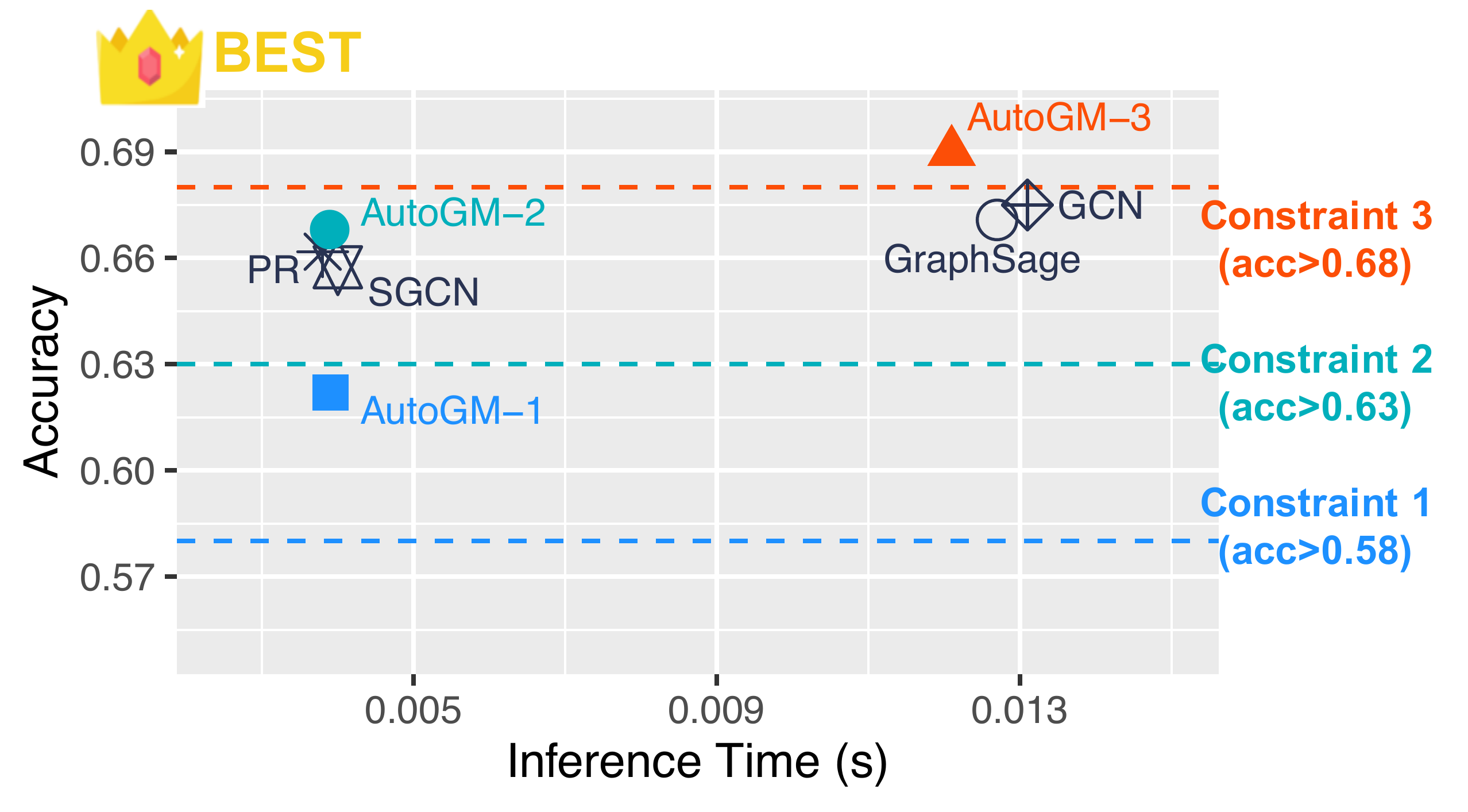}
	}
	\subfigure[Time constraints on the Citeseer dataset]
	{
		\label{fig:citeseer:time}
		\includegraphics[width=.48\linewidth]{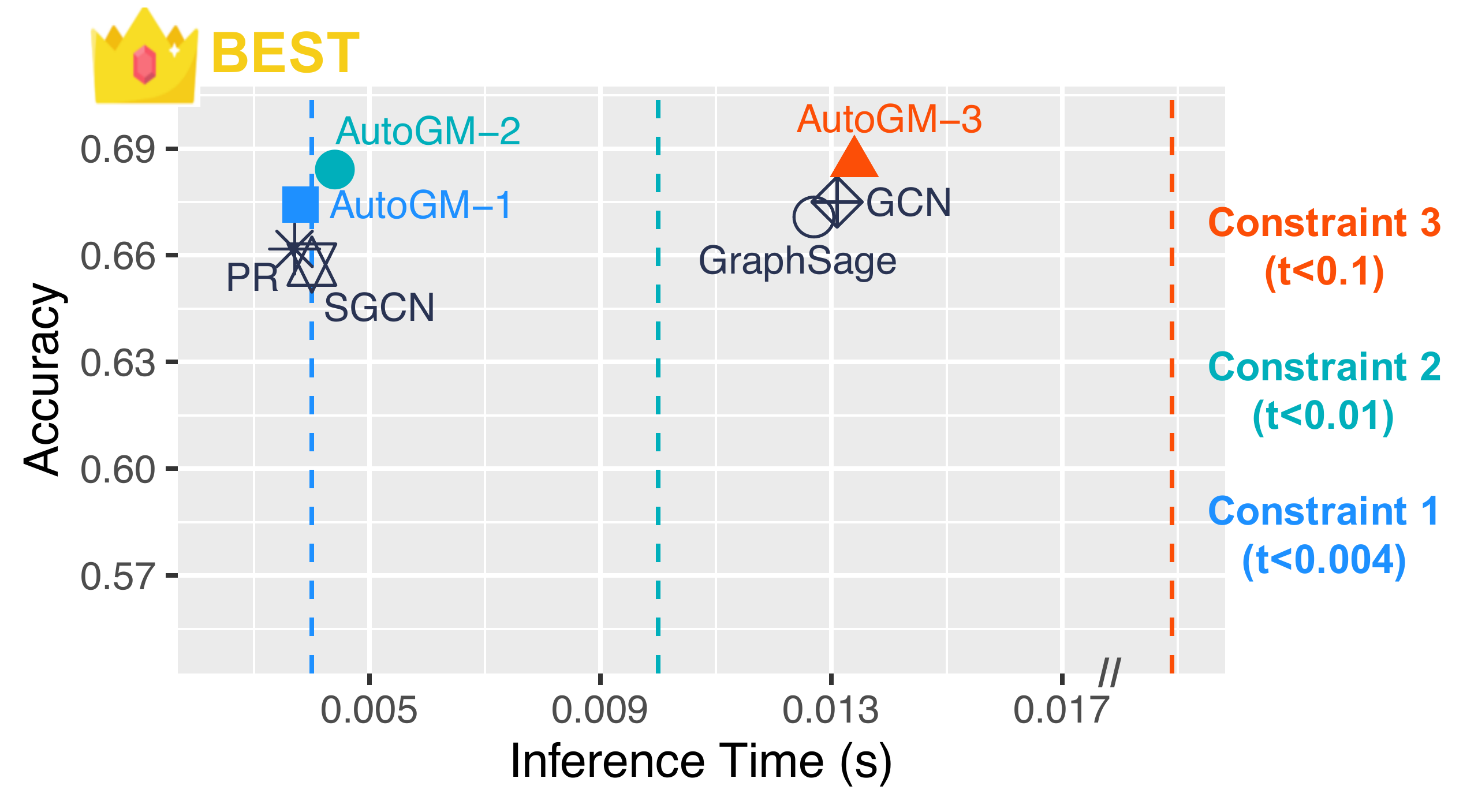}
	}
	\caption
	{ 
		\textbf{\method finds novel graph algorithms with the best accuracy/inference time trade-off}:
		(a) Given three accuracy lower bounds (i.e., $0.58, 0.63, 0.68$), \method generates three novel graph algorithms minimizing inference time.
		(b) Given three inference time upper bounds (i.e., $0.004, 0.01, 0.1$ seconds), \method generates three novel graph algorithms maximizing accuracy.
	}
	\label{fig:citeseer}
\end{figure*}

Hence, in this paper, we first unify various graph mining algorithms under our \frame framework, then propose an automated system for graph algorithm development, \method.
We target graph algorithms that pass messages --- propagate scores in the PageRank terminology ~\cite{kleinberg1999authoritative,page1999pagerank} --- along edges to summarize the graph structure into nodes statistics.
\frame manipulates five parameters of the message passing mechanism: the dimension of the communicated messages, the number of neighbors to communicate with (width), the number of steps to communicate for (length), the nonlinearity of the communication, and the message aggregation strategy. 
Different parameter settings yield novel graph algorithms, as well as existing algorithms, ranging from conventional graph mining algorithms (like PageRank) to graph neural networks.

\method leverages the parameter search space defined in \frame to address a practical problem: given a real-world scenario, what is the graph mining algorithm with the best speed/accuracy trade-off? 
In real-world scenarios, practitioners optimize performance under a computational budget~\cite{lemaire2019structured,li2019partial}.
\method defines a novel budget-aware objective function capturing the speed/accuracy trade-off, then maximizes the objective function to find the optimal parameter set of \frame, resulting in a novel graph mining algorithm tailored for the given scenario.

The goal of our work is to empower practitioners without much expertise in graph mining to deploy algorithms tailored to their specific scenarios.
The main contributions of this paper are as follows:

\begin{itemize}[leftmargin=10pt]
	\item {
		\textbf{Unification:}
		%\frame unifies various graph mining algorithms as instantiations of a message passing framework with five parameters: dimension, width, length, nonlinearity, and aggregation strategy. 
		\frame is a unified framework for message-passing based graph algorithms.
		\frame provides the parameter search space necessary to automate graph mining algorithm development.
	}
	\item {
		\textbf{Automation:}
		 \method is an automated system for graph mining algorithm development. 
		 Based on the search space defined by \frame, \method finds the optimal graph algorithm using Bayesian optimization.
	}
	\item {
		\textbf{Budget awareness:}
		\method maximizes accuracy of an algorithm under a computational time budget, or minimizes the computational time of an algorithm under a lower bound constraint on accuracy.
	}
	\item {
		\textbf{Effectiveness:}
		\method finds novel graph mining algorithms with the best speed/accuracy trade-off compared to existing models with heuristic parameters (Figure~\ref{fig:citeseer}).
	}
\end{itemize}
\noindent Table~\ref{tab:symbols} gives a list of symbols and definitions.

\noindent{\bf Reproducibility}: Our code is publicly available \codeurl.

\section{Background \& Related work}
\label{sec:related_works}
AutoML is the closest line of related work and the main inspiration for this paper.
AutoML algorithms are developed to automate the process of algorithm selection and hyperparameter optimization in the machine learning community.
%Machine learning (ML) algorithms are rarely parameter-free: we must often specify hyperparameters controlling the rate of learning or the capacity of the model.
%Applying an ML algorithm to a given scenario often requires tuning many hyperparameters --- a process that requires expert experience, brute-force search, and graduate student sweat. 
%AutoML is born out of the necessity to automatize this costly hyperparameter tuning process.
%AutoML formalizes hyperparameter tuning as the optimization of an unknown black-box function and invokes algorithms developed for such problems.
%The paradigm has been applied successfully to hyperparameter search (e.g., to tune learning rate, weight decay, dropout probability).
%Recently, AutoML has moved beyond low-dimensional hyperparameter optimization to problems with more sophisticated search spaces.
%The field that is most closely related to our approach is Neural Architecture Search (NAS), which focuses on the problem of searching for the deep neural network architecture with the best performance.
The most closely related to our work in AutoML is Neural Architecture Search (NAS), which focuses on the problem of searching for the deep neural network architecture with the best performance.
The search space includes the number of layers, the number of neurons, and the type of activation functions, among other design decisions. 
NAS broadly falls into three categories: evolutionary algorithms (EA), reinforcement learning (RL), and Bayesian optimization (BO).

EA-based NAS~\cite{floreano2008neuroevolution, kitano1990designing, liu2017hierarchical} explores the space of architectures by making a sequence of changes (inspired by evolutionary mutations) to networks that have already been evaluated.
In RL-based NAS~\cite{zhong2017practical,zoph2016neural}, a recurrent neural network iteratively decides if and how to extend a neural architecture; the non-differentiable cost function is optimized with stochastic gradient techniques borrowed from the RL literature.
Finally, BO-based NAS~\cite{kandasamy2018neural} models the cost function probabilistically and carefully determines future evaluations to minimize the total number of evaluated architectures.
%EA and RL are more flexible than BO in the specification of the search space, but BO is often the most efficient search technique if it is compatible with the search space.
Since EA and RL-based NAS need to evaluate a vast number of architectures to find the optimum, these approaches are not ideally suited for neural architecture search~\cite{kandasamy2018neural}.
% which are expensive to evaluate~\cite{kandasamy2018neural}. 
On the other hand, BO emphasizes being cautious in selecting which architecture to try next to minimize the number of evaluations.
As we discuss later, this makes BO suitable for our problem.
In the following section, we give a brief description of Bayesian Optimization.
%Most existing autonomous architecture search approaches focus only on high performance.
%But real-world systems often have computation constraints, leading to low efficiency when deploying popular high accuracy models ~\cite{he2016deep}.
%A more principled approach is to select the network that offers the best speed/accuracy trade-off on a target platform.
%Partial Order Pruning algorithm~\cite{li2019partial} prunes the search space to concentrate on architectures that are more likely to lift the boundary of the speed/accuracy trade-off.

\subsection{Bayesian Optimization}
\label{sec:related_works:BO}
Given a black-box objective function $f$ with domain $\X$, BO sequentially updates a Gaussian Process prior over $f$.
At time $t$, it incorporates results of previous evaluations $1, . . . , t-1$ into a posterior $P(f|\Data_{1:t-1})$ where $\Data_{1:t-1}=\{x_{1:t-1}, f(x_{1:t-1})\}$.
BO uses this posterior to construct an acquisition function $\phi_t(x)$ that is an approximate measure of evaluating $f(x)$ at time $t$.
BO evaluates $f$ at the maximizer of the acquisition function $x_t= \arg \max_{x\in\X} \phi_t(x)$.
The evaluation $f(x_t)$ is then incorporated into the posterior $P(f|\Data_{1:t})$, and the process is iterated.

The evaluation point $x_t$ chosen by the acquisition function is an approximation of the maximizer of $f$.
After $T$ iterations, BO returns the parameter set of the maximum $f$ among $x_{1:T}$.
When choosing the point $x_t$ to evaluate, the acquisition function $\phi_t(x)$ trades off exploration (sampling from areas of high uncertainty) with exploitation (sampling areas likely to offer an improvement over the current best observation). 
This cautious trade-off helps to minimize the number of evaluations of $f$. 
More details about BO can be found in~\cite{brochu2010tutorial}.

However, these AutoML techniques cannot be directly applied to graph mining, as they require first formalizing autonomous algorithm selection as an optimization problem in a hyperparameter search space. 
However, before \frame, the hyperparameter search space for graph mining was not even defined due to the lack of unity among algorithms.
Hence, our proposed \frame and \method allow the graph mining field to exploit state-of-the-art techniques developed in AutoML.

\section{Unified Graph Mining Framework}
\label{sec:frame}
\begin{table}[!t]
	\vspace{-5mm}
	\centering
	\caption{Commonly used notation.}
	\begin{tabular}{cl}
		\toprule
		\textbf{Symbol} & \textbf{Definition} \\
		\midrule
		$G$ & {input graph} \\
		$n,m$ & {numbers of nodes and edges in $G$}\\
		$\A$ & ($n \times n$) binary adjacency matrix of $G$\\
		$d_0$ & {dimension of input feature vectors}\\
		$d$& {dimension of communicated messages}\\
		$k$& {number of message passing steps}\\
		$w$& {number of neighbors sampled per node} \\
		$l$ & {binary indicator for nonlinearity}\\
		$a$ & {categorical aggregation strategy}\\
		$X_0$ & {($n \times d_0$) input feature vectors}\\
		$X_i$& {($n \times d$) $i$-th layer message vectors $(i=1\dots k)$}\\
		$W_1$ & {($d_0 \times d$) 1st layer transformation matrix}\\
		$W_i$ & {($d \times d$) $i$-th layer transformation matrix} \\
		& {$(i=2\dots k)$}\\
		$\phi(x)$ & $\begin{cases} ReLU(x)~\text{if}~l = \text{True} \\ x~\text{otherwise} \end{cases}$\\
		\bottomrule
	\end{tabular}
	\label{tab:symbols}
\end{table}

In this section, we first motivate the message passing scheme (Section~\ref{sec:frame:mp}). 
We then propose our unified framework \frame (Section~\ref{sec:frame:algorithm}), explain how existing algorithms fit in the framework (Section ~\ref{sec:frame:reproduction}), and further analyze how \frame bridges the conceptual gap between conventional graph mining and graph neural networks (Section~\ref{sec:frame:analysis}).
Finally, we outline how to choose parameters of \frame given a specific scenario (Section~\ref{sec:frame:parameter}). 

\subsection{Message Passing}
\label{sec:frame:mp}

A goal common to many graph mining algorithms is to answer queries at the node level (e.g., node clustering, classification, or recommendation) based on global graph information (e.g., edge structure and feature information from other nodes). 
To transmit the information necessary to answer such queries, in classical graph mining algorithms, nodes \textit{propagate scalar scores to} their neighbors, while in graph neural networks, nodes \textit{aggregate feature vectors from} their neighbors. 
In short, both families of algorithms pass messages among neighbors: scalars or vectors, inbound or outbound. 
The intuition behind these message passing algorithms is that
whatever the task at hand, connectivity/locality matters: connected/nearby nodes are more similar (clustering), informative (classification), or relevant (recommendation) to each other than disconnected/distant nodes.
Our unified framework targets graph algorithms that use the message passing mechanism.

\subsection{\frame}
\label{sec:frame:algorithm}

\begin{figure}[!t]
	\vspace{-5mm}
	\centering
	\includegraphics[width=.95\linewidth]{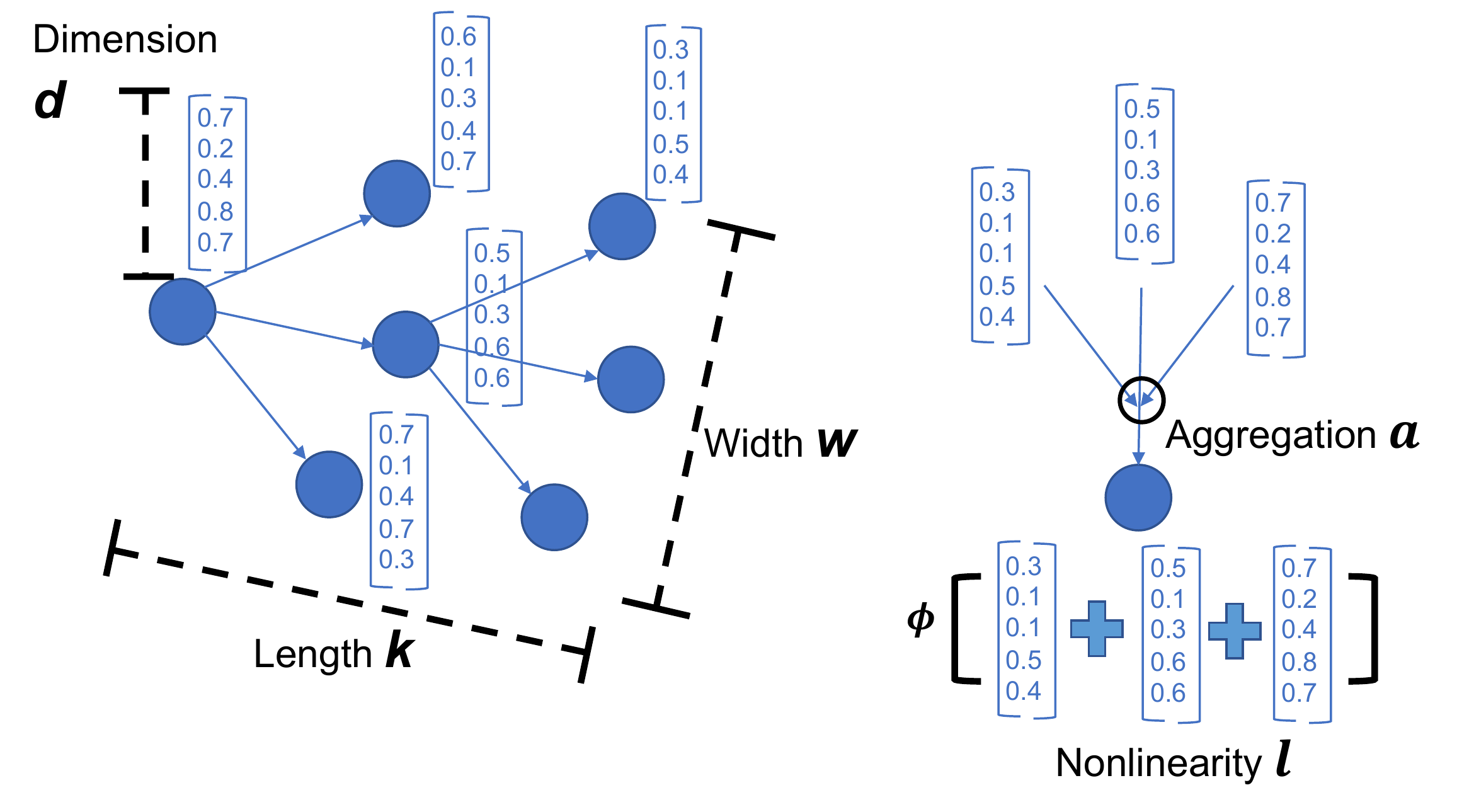}
	\caption
	{ 
		\frame defines the message passing mechanism based on five parameters: the dimension $d$, length $k$, width $w$, nonlinearity $l$, and aggregation strategy $a$.
	}
	\label{fig:overview}
\end{figure}

We propose a unified framework \frame for graph mining algorithms that employ the message passing scheme. 
\frame defines the message passing mechanism based on five parameters: 
\begin{itemize}[leftmargin=10pt]
	\item {
		\textbf{Dimension $d \in \Z_{>0}$} of passed messages. If $d = 1$, messages are scalar scores, otherwise they are $d$-dimensional embedding vectors. 
	}
	\item {
		\textbf{Width $w \in {\Z} \cup \{-1\}$} decides the number of neighbors each node communicates with. If $w = -1$, nodes communicate with all their neighbors.
	}
	\item {
		 \textbf{Length $k \in {\Z}$} decides the number of message passing steps.
	}
	\item {
		\textbf{Nonlinearity} $l \in \{\text{True}, \text{False}\}$ decides whether to use nonlinearities in the message passing or not.
	}
	\item {
		\textbf{Aggregation strategy} $a$ decides if a node sends a message to itself and how to normalize the sum of incoming messages.
	}
\end{itemize}
\noindent Figure~\ref{fig:overview} shows how each parameter regulates message passing under \frame.

The input of \frame is a graph $G = (V, E)$ and a matrix $X_0$ of size $(n \times d_0)$ containing $d_0$-dimensional initial node statistics for all $n$ nodes --- either scalar scores or feature vectors. Note that $d_0$ could be different from $d$, the dimension of the passed messages. The output of \frame is a set of $d$-dimensional node embeddings. These embeddings contain information from the node's neighborhood and can be exploited in an output layer which is specialized to a given application (e.g., a logistic regression for node classification.)

Algorithm~\ref{alg:frame} outlines how \frame passes messages across a graph based on a set of five parameters ($d, k, w, l, a$). 
\frame first initializes node statistics (line~\ref{alg:frame:init}), then iteratively passes messages among neighboring nodes $k$ times.
In the $i$-th message passing step, \frame randomly samples $w$ neighbors to communicate with for each node (line~\ref{alg:frame:samp}) and aggregates messages from sampled neighbors with a strategy decided by the parameter $a$ (lines~\ref{alg:frame:mat-agg} and~\ref{alg:frame:agg}).
Then \frame transforms the aggregated messages linearly with a matrix $W_{i}$ (line~\ref{alg:frame:trans}) and finally passes them through a function $\phi$ decided by the parameter $l$ (line~\ref{alg:frame:nonlinear}).

Let us explain in further detail the neighbor sampling and message aggregation steps.
Neighbor sampling (line ~\ref{alg:frame:samp}) can be expressed as generating a matrix $A_\text{samp} = \textit{Sample}(A)$ by randomly zeroing out entries of the binary adjacency matrix $A$.
Message aggregation (lines~\ref{alg:frame:mat-agg} and \ref{alg:frame:agg}) is defined by the aggregation strategy $a \in$ \{SA, SS, SN, NA, NS, NN\}.
The first letter in \{S, N\} determines whether a node sends a message to itself or not (Self-loop or No-self-loop). The second letter in \{A, S, N\} determines how to normalize the sum of incoming messages (Asymmetric, Symmetric, or No-normalization).
Each aggregation strategy $a$ results in an aggregation matrix $A_\text{agg} = \textit{Aggregate}(A_\text{samp})$, explained in Table ~\ref{tab:aggregation}.
Multiplying messages $X_{i-1}$ from the previous step by the matrix $A_\text{agg}$ corresponds to aggregating messages from neighboring nodes.
 
Letting $f_i$ denote the $i$th layer of message passing, we can summarize \frame as follows:
\begin{align*}
A_{\text{samp}} &= \textit{Sample}(A) \\
A_{\text{agg}} &= \textit{Aggregate}(A_{\text{samp}}) \\
X_k 
&= f_{k}(X_{k-1}) = \phi(A_{\text{agg}} X_{k-1} W_k) \\
&=  f_{k}(f_{k-1}(\dots f_{1}(X_0)))
\end{align*}
\noindent 
$X_0$ is the ($n \times d_0$) matrix of initial statistic vectors, $X_i$ is the ($n \times d$) matrix of statistic vectors at step $i$ for $(i=1\dots k)$.
$W_1$ and $W_i$ are ($d_0 \times d$) and ($d \times d$) transformation matrices respectively $(i=2\dots k)$.

\begin{algorithm}[t]
	\small
	\begin{algorithmic}[1]
		\caption{\frame Algorithm}
		\label{alg:frame}
		\REQUIRE initial node statistics $X_0$, binary adjacency matrix $\A$
		\ENSURE node embeddings $X_k$
		\STATE Initialize node statistics $X_0$ \label{alg:frame:init}
		\FOR{message passing step $i = 1$; $i \le k$; $i$++}
		\STATE Sample neighbors for each node: $A_{\text{samp}} \gets \textit{Sample}(A)$ \label{alg:frame:samp}
		\STATE Generate aggregation matrix: $A_{\text{agg}} \gets \textit{Aggregate}(A_{\text{samp}})$ \label{alg:frame:mat-agg}
		\STATE Aggregate messages $X_i \gets A_\text{agg} X_{i-1}$ \label{alg:frame:agg}
		\STATE Multiply with transformation matrix: $X_i \gets X_i W_i $\label{alg:frame:trans}
		\STATE Pass through nonlinear function: 
		$X_i \gets \phi(X_i) = \begin{cases} ReLU(X_i)~\text{if}~l = \text{True} \\ X_i~\text{otherwise} \end{cases}$\label{alg:frame:nonlinear}
		\ENDFOR 
		\BlankLine
		\RETURN $X_k$
	\end{algorithmic}
\end{algorithm}

\subsection{Reproduction of Existing Algorithms}
\label{sec:frame:reproduction}

In this section, we introduce the most popular graph mining algorithms exploiting the message passing scheme and show how they can be presented under \frame.
Table~\ref{tab:algorithms} shows how to set initial node statistics and parameters $(d, k, w, l, a)$ of \frame to reproduce the original graph algorithms.

\textbf{PageRank}~\cite{page1999pagerank} scores nodes in a graph based on their global relevance/importance, and was initially used by Google for webpage recommendation.
PageRank initializes all $n$ nodes in the graph with a score of $\frac{1}{n}$. 
Then, every node iteratively propagates its score across the graph with a decay coefficient $0 < c < 1$ to ensure convergence.
Under \frame, PageRank propagates scalar scores ($d=1$) to all neighbors ($w=-1$) with no nonlinear unit ($l= \text{False}$) until scores have converged ($k=\infty$), and aggregates messages with no self-loop and asymmetric normalization ($a =$ NA). 
Note that the $(d \times d)$ transformation matrix $W$ in \frame becomes a scalar value and corresponds to the decay coefficient $c$.

\textbf{Personalized PageRank} (PPR)~\cite{bahmani2010fast} and \textbf{Random Walk with Restart} (RWR)~\cite{yoon2018fast,yoon2018tpa} build on PageRank to estimate the relevance of nodes in the perspective of a specific set of seed nodes thus enable personalized recommendation. 
%RWR~\cite{yoon2018tpa} initializes a specific set of source nodes with positive scores and other nodes with zero scores, then propagates scores with a decay coefficient until convergence. 
%The final score of a node reflects its relevance for source nodes based on an assumption of homophily (connectivity implies similarity/relevance).
Under \frame, the only difference of PPR/RWR from PageRank is the initial node scores: RWR/PPR place varying positive scores on the set of seed nodes and zero scores on others.
PPR/RWR have the same set of $(d, k, w, l, a)$ as PageRank.

\textbf{Pixie}~\cite{eksombatchai2018pixie}, introduced by Pinterest, complements the ideas of PPR and RWR with neighbor sampling to deal with billions of nodes in real-time.
Pixie fixes the number of message passing operations and stays within a computation budget.
To reproduce this under \frame, Pixie fixes the product of $k$ and $w$ to a constant number (e.g., $2,000$ from~\cite{eksombatchai2018pixie}):
after $k$ is sampled, $w$ is decided as $\frac{2,000}{k}$.
Pixie has the same initial node statistics and parameter $d=1, l=\text{False}$, and $a=\text{NA}$ with PPR/RWR. 
%For the parameter $a$, Pixie sums up scores from the sampled neighbors (NN: no self-loop and no normalization).

\begin{table}
	\centering
	\caption{The aggregation strategy $a$ decides if a node sends a message to itself or not (Self-loop or No-self-loop) and how to normalize the sum of incoming messages (Asymmetric, Symmetric, or No-normalization). Each combination corresponds to an aggregation matrix $A_\text{agg} = \textit{Aggregate}(A)$ in the table. Notation: $n$ is the number of nodes in a graph, $\A$ is a ($n\times n$) binary adjacency matrix, $\D$ is a ($n\times n$) diagonal matrix where $\D_{ii} = \sum_{j}\A_{ij}$, and $\I_n$ is an identity matrix of size $n$.}
	\begin{tabular}{l|ll}\hline
		\toprule
		& Self-loop (S) & No-self-loop (N) \\
		\midrule
		Asymmetric (A) & $D^{-1} (A + I_n)$ & $D^{-1} A$\\
		Symmetric (S) & $D^{-1/2} (A + I_n) D^{-1/2}$ & $D^{-1/2} A D^{-1/2}$ \\
		No-normalization (N) & $(A + I_n)$ & $A$ \\
		\bottomrule
	\end{tabular}
	\label{tab:aggregation}
\end{table}

\begin{table*}[t]
	\vspace{-2mm}
	\centering
	\caption{Graph mining algorithms can be fully reproduced under \frame with the respective initial node statistics and parameters ($d, k, w, l, a$).
	Notation: $n$ is the number of nodes, $\A$ denotes an ($n\times n$) binary adjacency matrix, $\D$ denotes an ($n\times n$) diagonal matrix where $\D_{ii} = \sum_{j}\A_{ij}$, $\I_n$ denotes an identity matrix of size $n$, $\textsc{N}(u)$ denotes the set of sampled neighbors of node $u$, and $0 <c < 1$ is a decay coefficient. 
	For PageRank, see the formulation given in~\cite{yoon2018tpa}.
	}
	\begin{tabular}{l|l|l|lllll}
		\toprule
		\textbf{Algorithm} & Original message passing equation & Initial node statistics & $d$ & $k$ & $w$ & $l$ & $a$ \\ \midrule
		PageRank~\cite{page1999pagerank} & $X_k= c (\D^{-1}\A) X_{k-1}$ & $\frac{1}{n}$ for all nodes &  1  & $\infty$  & -1  & False  & NA \\
		Pixie~\cite{eksombatchai2018pixie} & $X_k(u) = \sum_{v\in\textsc{N}(u)}X_{k-1}(v)$ & 1 for seeds, 0 others & 1 & sample & $\frac{2000}{k}$ & False  & NA \\
		GCN~\cite{kipf2016semi} & $X_k = ReLU\left((\D^{-\frac{1}{2}}(\A + \I_n)\D^{-\frac{1}{2}}) X_{k-1} W_{k}\right)$ & feature vectors & 64 & 2 & -1 & True & SS \\
		GraphSAGE~\cite{hamilton2017inductive} & $X_k(u) = ReLU \left(\frac{1}{|\textsc{N}(u)| + 1} \sum_{v\in\textsc{N}(u) \cup u} X_{k-1}(v) W_{k}\right)$ & feature vectors & 64 & 2 & 25 & True & SA \\
		SGCN~\cite{wu2019simplifying} & $X_k = \D^{-\frac{1}{2}}(\A + \I_n)\D^{-\frac{1}{2}} X_{k-1} W_{k}$ & feature vectors & 64  & 2 & -1 & False & SS \\ \bottomrule                                                   
	\end{tabular}
	\label{tab:algorithms}
\end{table*}

\textbf{Graph Convolutional Networks} (GCNs)~\cite{kipf2016semi} are a variant of Convolutional Neural Networks that operates directly on graphs.
GCNs stack layers of first-order spectral filters followed by a nonlinear activation function to learn node embeddings.
Under \frame, given node feature vectors as initial node statistics, GCN passes message vectors ($d=64$) to all neighbors ($w=-1$) with nonlinear units ($l = \text{True}$) across two-layered networks ($k=2$) and aggregates messages with a self-loop and symmetric normalization ($a =$ SS).

\textbf{GraphSAGE}~\cite{hamilton2017inductive} extends GCN with neighbor sampling. 
GraphSage with a mean aggregator averages statistics of a node and its sampled neighbors. 
Under \frame, GraphSAGE-mean has the same parameters as GCN except $w$ and $a$.
GraphSAGE-mean samples a fixed number of neighbors to communicate with ($w=25$) and normalizes the aggregated messages asymmetrically ($a=$ SA).

\textbf{Simplified GCN} (SGCN)~\cite{wu2019simplifying} reduces the excess complexity of GCN by removing the nonlinearities between GCN layers and collapsing the resulting function into a single linear transformation.
With fewer parameters to train, SGCN is computationally more efficient than GCN but shows comparable performance on various tasks.
Under \frame, SGCN has the same parameters with GCN except $l$.
SGCN does not use any nonlinear unit ($l = \text{False}$).

Table~\ref{tab:algorithms} presents the original message passing equations of the existing graph algorithms.
Those equations can be fully reproduced from Algorithm~\ref{alg:frame} with the proper inital node statistics and parameter sets listed in Table~\ref{tab:algorithms}.

\subsection{Conventional GM vs. GNNs}
\label{sec:frame:analysis}

As shown, conventional graph algorithms (e.g., PPR, RWR, Pixie) and recent GNNs are unified under \frame.
However, before this work, these algorithms were not analyzed in the same framework.
What has prevented them from being combined?
Two main differences --- the use of node feature information and trainability --- are the culprits.
While GNNs exploit additional node feature information and labels with semi-supervised learning, conventional graph algorithms do not.
We analyze this apparent gap and show how \frame reconciles both families of algorithms.

\textbf{Node feature information:}
Conventional graph algorithms do not exploit node features, but instead, choose a set of seed nodes to initialize with scores suitable for a given application.
%Under \frame, these families of algorithms \note{are reproducible} with distinct initializations for node statistics.
Under \frame, these algorithms are also applicable with node features by maintaining the same values for parameters $(d=1, k, w, l, a)$, but setting initial input dimension $d_0$ to be the input feature dimension and using a $1$st layer tranformation matrix $W_1$ of size $(d_0 \times 1)$. 
This would yield a new version of PageRank or PPR that exploits feature information.

\textbf{Semi-supervised learning:} 
In GNNs, the transformation matrix $W$ is trained with semi-supervised learning using node labels.
On the other hand, conventional graph algorithms do not have a training phase in advance of an inference phase.
However, conventional algorithms are trainable: 
the decay coefficient $c$ in PageRank, PPR, and RWR corresponds to an $(1 \times 1)$ transformation matrix $W$ under \frame.
Because of its low dimension, the $(1 \times 1)$ transformation matrix could be set heuristically (e.g., $c=0.85$ in PageRank).
But we could use label information to train this $(1 \times 1)$ matrix $W$ with gradient descent as we train it in GNNs. 

In our experiments, we show how to train conventional algorithms (PageRank and Pixie) with feature information. 

\subsection{Parameter Selection}
\label{sec:frame:parameter}

We explain the effects of parameters ($d, k, w, l, a$) on the performance of graph algorithms and how to choose the proper parameters by illustrating the existing algorithm design. 

\begin{itemize}[leftmargin=10pt]
		\item {\textbf{Dimension $d$:} 
		High dimensions of messages enrich the expressiveness of graph algorithms by sacrificing speed.
		If an application prioritizes fast and simple algorithms, scalar messages (e.g., $d=1$ in Pixie) are suitable.
		In contrast, when applications prioritize rich expressiveness of messages and accuracy, high dimensional vectors (e.g., $d = 64$ in GNNs) are more appropriate.
		}
		\item {\textbf{Length $k$:} 
		$k$ decides the size of neighborhood where a graph algorithm assumes locality --- where nearby nodes are considered informative.
		For instance, GCNs assume that a small neighborhood is relevant ($k = 2$).
		However, when there are label sparsity issues, GNNs propagate toward large scopes ($k=7$) to transmit label information from distant nodes.
		Large $k$ results in a long computation time but does not guarantee a high accuracy.
		}
		\item {\textbf{Width $w$:} 
		Large $w$ lets algorithms aggregate information from more neighbors, leading to a possible increase in accuracy.
		At the same time, large $w$ requires more message passing operations, resulting in longer computation time.
		In graphs with billions of nodes, like the Pinterest social network, small $w$ is necessary to answer queries in real-time (as done by Pixie).
		}
		\item {\textbf{Nonlinearity $l$:} 
		Nonlinearities enhance the expressiveness of graph algorithms at the cost of speed. They are suitable for anomaly detection systems that require high accuracy (e.g., GNNs for infection detection in medical applications). 
		In contrast, omitting nonlinearity is appropriate for fast recommender systems in social networks (e.g., Pixie in Pinterest).
		}
		\item {
		\textbf{Aggregation strategy $a$:}
		The self-loop decides whether a node processes its own embedding during message passing. 
		GNNs include a self-loop to complement a node's features with information from its neighborhood.
		Conversely, PageRank and RWR do not include a self-loop as they want to spread information from a source node to the rest of the graph to figure out the graph structure.
		Normalization prevents numerical instabilities and exploding/vanishing gradients in GNNs. 		
		}
\end{itemize}

\noindent In our experiments, we explore how the five parameters affect the performance of graph algorithms empirically.

\section{Automation of Graph Mining Algorithm Development}
\label{sec:method}
With the proper parameter selection, \frame could output a graph algorithm tailored for a specific application.
However, the parameter selection process still relies on the intuition and domain knowledge of practitioners, which would prevent non-experts in graph mining from fully exploiting \frame.
How can we empower practitioners without much expertise to deploy customized algorithms?
We introduce \method, which generates an optimal graph algorithm autonomously given a user's scenario.

When designing an algorithm for an application, we need to consider two primary metrics: computation time and accuracy, which usually trade off each other.
Take, for example, a developer who aims to develop an online recommender system that makes personalized recommendations to a large number of users at the same time.
At first, she employs a state-of-the-art GNN model (in terms of accuracy) but finds that the computation time is too long for her application.
Then the developer seeks an alternative simple graph algorithm that runs faster than a time budget by sacrificing accuracy.
\method incorporates this practical issue of finding the best speed-accuracy trade-off into the graph algorithm generation problem.
\method answers two questions: 
1) given the maximum acceptable computation time, which graph algorithm maximizes accuracy? 
2) given minimum accuracy requirements, which graph algorithm minimizes computation time?

We first formalize our budget-aware graph algorithm generation problem as a constrained optimization problem.
Then we replace the constrained problem with an unconstrained optimization problem using barrier methods (Section~\ref{sec:method:objective}).
We explain why Bayesian optimization is well-suited for this unconstrained problem (Section~\ref{sec:method:bo}).
Finally, we describe how \method solves the optimization problem using Bayesian optimization (Section~\ref{sec:method:algorithm}).

\subsection{Budget-aware objective function}
\label{sec:method:objective}

Letting $x$ denote a graph algorithm, $g(x)$ and $h(x)$ indicate the computation time and accuracy of $x$, respectively.
Then an optimal graph algorithm generation problem with an accuracy lower bound $h_{\min}$ is presented as a constrained optimization as follows:
\begin{align}
	{x}_{opt} = \mbox{argmin}_{x}~g(x)~\mbox{subject to}~h(x) - h_{\min}\ge 0 
\label{equ:constrained}
\end{align}
One of the common ways to solve a constrained optimization problem is using a barrier method~\cite{vanderbei2001linear}, replacing inequality constraints by a penalizing term in the objective function.
We re-formulate the original constrained problem in Equation~\ref{equ:constrained} as an equivalent unconstrained problem as follows:
\begin{align}
	{x}_{opt} =\mbox{argmin}_x~g(x) + I_{h(x)-h_{\min}\ge0}(x) 
\label{equ:unconstrained}
\end{align}
where the indicator function $I_{h(x)-h_{\min}\ge0}(x) = 0$ if $h(x)-h_{\min} \geq 0$ and $\infty$ if the constraint is violated.
Equation~\ref{equ:unconstrained} eliminates the inequality constraints, but introduces a discontinuous objective function, which is challenging to optimize.
Thus we approximate the discontinuous indicator function with an optimization-friendly log barrier function.
The log barrier function, defined as $-\log(h(x)-h_{\min})$ is a continuous function whose value on a point increases to infinity ($-\log0$) as the point approaches the boundary $h(x)-h_{\min} = 0$ of the feasible region.
Replacing the indicator function with the log barrier function yields the following optimization problem:
\begin{align}
	\f(x)&= g(x) - \lambda \log (h(x)-h_{\min}) \\
	{x}_{opt} &=\mbox{argmin}_x~ \f(x)
\label{equ:log_barrier}
\end{align}
\noindent 
$\f$ is our novel budget-aware objective function and $\lambda > 0$ is a penalty coefficient.
Equation~\ref{equ:log_barrier} is not equivalent to our original optimization problem, Equation~\ref{equ:constrained}.
However, as $\lambda$ approaches zero, it becomes an ever-better approximation (i.e., $-\lambda \log (h(x)-h_{\min})$ approaches $I_{h(x)-h_{\min}\ge0}(x)$)~\cite{vanderbei2001linear}. 
The solution of Equation~\ref{equ:log_barrier} ideally converges to the solution of the original constrained problem. 
Now, our budget-aware graph algorithm generation problem is formulated as a minimization problem of $\f$.

Given a minimum accuracy constraint $acc_{\min}$, we set $g(x) = time$ to minimize and $h(x) - h_{\min}= acc - acc_{\min} \geq 0$ as a constraint. 
On the other hand, given a maximum inference time constraint $time_{\max}$, we want to maximize accuracy while observing the time constraint.
Then we set $g(x) = -acc$ to minimize and  $h(x) - h_{\min}= time_{\max} - time \geq 0$ as a constraint.

\begin{algorithm} [t!]
	\small
	\begin{algorithmic}[1]
		\caption{\method Algorithm}
		\label{alg:method}
		\REQUIRE minimum accuracy (or maximum inference time) constraint, target dataset, BO search budget
		\ENSURE a graph algorithm (i.e. five parameters of \frame)
		\FOR{iteration $i = 1$; $i<$ BO search budget; $i$++}
		\STATE Choose a point $(d, k, w, l, a)$ to evaluate \label{alg:method:choose}
		\STATE Generate a graph mining algorithm $A$ from $(d, k, w, l, a)$ \label{alg:method:algo}
		\STATE Train $A$ on the training set \label{alg:method:train}
		\STATE Evaluate $A$ and measure $acc$, $time$ on the validation set \label{alg:method:evaluate}
		\STATE Evaluate $\f(acc, time)$ and update posterior of $\f$ \label{alg:method:update}
		\ENDFOR 
		\RETURN a parameter set with the minimum $\f$
	\end{algorithmic}
\end{algorithm}

\subsection{Bayesian optimization}
\label{sec:method:bo}

Under \frame, a graph algorithm $x$ is defined by a set of parameters $(d, k, w, l, a)$.
Then search space $\X$ for the optimization problem becomes a five-dimensional space of parameters $(d, k, w, l, a)$.
Suppose we set cardinalities for each parameter as $300, 30, 50, 2,$ and $6$, respectively (i.e., $0<d\in\mathbb{Z}\le300, 0<k\in\mathbb{Z}\le30, 0<w\in\mathbb{Z}\le50, l\in\{True,False\}, a\in\{NA,NS,NN,SA,SS,SN\}$).
Then the number of unique architectures within our search space is $300\times30\times50\times2\times6 = 5.4\times 10^{6}$, which is quite overwhelming.
Moreover, training and validating a graph algorithm, especially on large datasets, takes significant time.
Thus it is impractical to search the space $\X$ exhaustively.
Most importantly, even if we could measure the computation time and accuracy ($g(x)$ and $h(x)$) of a graph algorithm and calculate the objective function $\f(x) = g(x) - \lambda \log (h(x)-h_{\min})$, we do not know the exact closed-form of $\f(x)=\f(d,k,w,l,a)$ in terms of the parameters $(d, k, w, l, a)$ nor its derivatives.
Thus, we cannot exploit classical optimization techniques that use derivative information.
To cope with these problems --- expensive evaluation and no closed-form expression nor derivatives --- which optimization technique is appropriate? 

Bayesian optimization (BO)~\cite{brochu2010tutorial} is the most widely-used approach to find the global optimum of a black-box cost function --- a function that we can evaluate but for which we do not have a closed-form expression or derivatives. 
Also, BO is cost-efficient with as few expensive evaluations as possible (more details in Section~\ref{sec:related_works:BO}). 
Therefore BO is well-suited to our problem to find the best parameter set $(d, k, w, l, a)$ given the expensive black-box objective function $\f(x)$.

\subsection{\method}
\label{sec:method:algorithm}

Users supply three inputs to \method: 
1) a budget constraint (the minimum accuracy or maximum computation time), 
2) a target dataset on which they want an optimized algorithm --- containing a graph, initial node scores, and labels for supervised learning --- and 
3) a search budget for Bayesian Optimization.
The search budget is given as the total number of evaluations in BO.
Then \method outputs the optimal graph mining algorithm (i.e., parameter set of \frame).

Algorithm~\ref{alg:method} outlines how \method works.
Until it has exhausted its search budget, \method repeats the process:
1) Pick a point $x = (d, k, w, l, a) \in \X$ to evaluate using an acquisition function of BO (line \ref{alg:method:choose}) then generate a graph algorithm $A$ from parameters $(d, k, w, l, a)$ (line \ref{alg:method:algo}).
2) Train $A$ on the training set (line \ref{alg:method:train}) and measure accuracy and inference time of $A$ on the validation set (line \ref{alg:method:evaluate}).
3) Evaluate the objective function $\f$ given the accuracy and inference time of $A$, then update a posterior model for $\f$ in BO (line \ref{alg:method:update}).
After all iterations, \method returns the parameter set $x = (d, k, w, l, a)$ with the minimum $\f$ among the evaluated points.
%Given a maximum acceptable inference time/a minimum accuracy requirement, \method returns the graph algorithm with the highest accuracy/the shortest inference time, respectively.

The search space of \method is not affected by the input but fixed to a five-dimensional space of parameters $(d, k, w, l, a)$.
The search time of \method is determined by the BO search budget (total number of evaluations) and evaluation time.
Since the evaluation time of a graph algorithm is often proportional to the input dataset's size, the total search time of \method is decided by the dataset.
%BO's minimization of the number of evaluations is especially crucial for large datasets, where other search methods would be intractable.
BO's minimization of the number of evaluations is especially efficient for large datasets which result in the long evaluation time.
Our main contribution is defining the graph algorithm generation problem as an optimization problem on a novel search space. 

\section{Experiments}
\label{sec:experiments}

\begin{figure*}[t!]
	\vspace{-5mm}
	\centering
	\subfigure[Accuracy constraints on the Cora dataset]
	{
		\label{fig:cora:acc}
		\includegraphics[width=.45\linewidth]{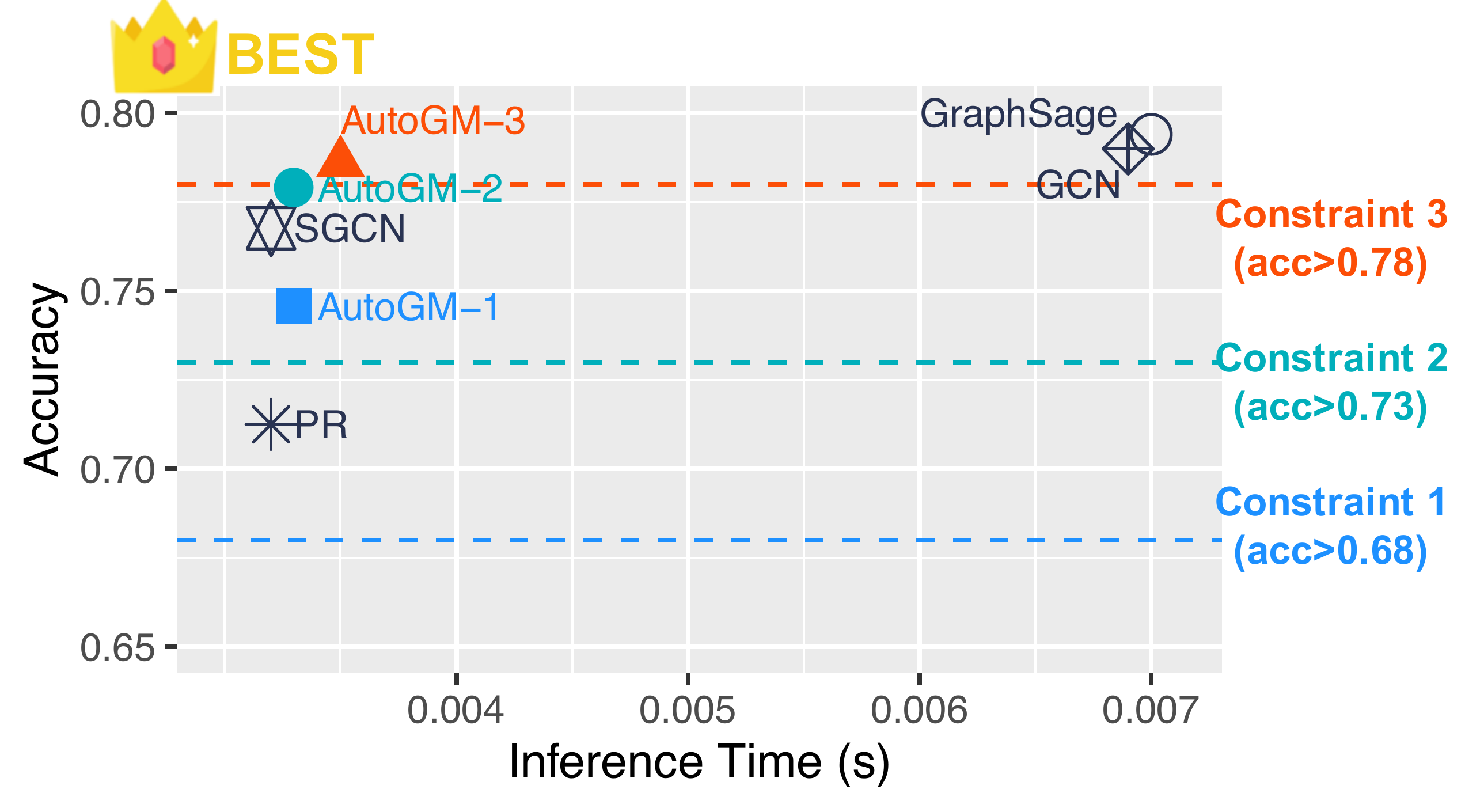}
	}
	\subfigure[Time constraints on the Cora dataset]
	{
		\label{fig:cora:time}
		\includegraphics[width=.45\linewidth]{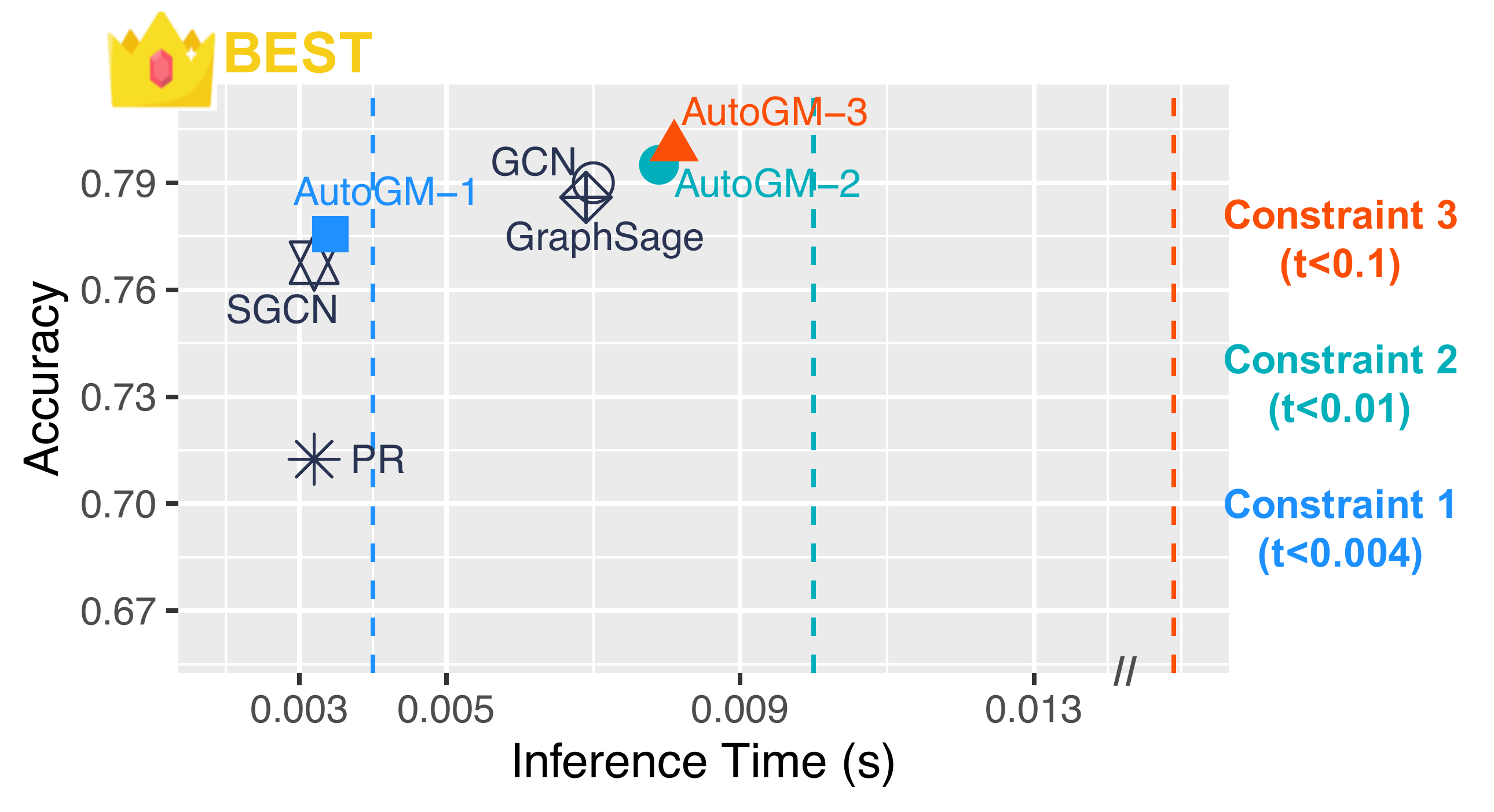}
	}
	\subfigure[Accuracy constraints on the Pubmed dataset]
	{
		\label{fig:pubmed:acc}
		\includegraphics[width=.45\linewidth]{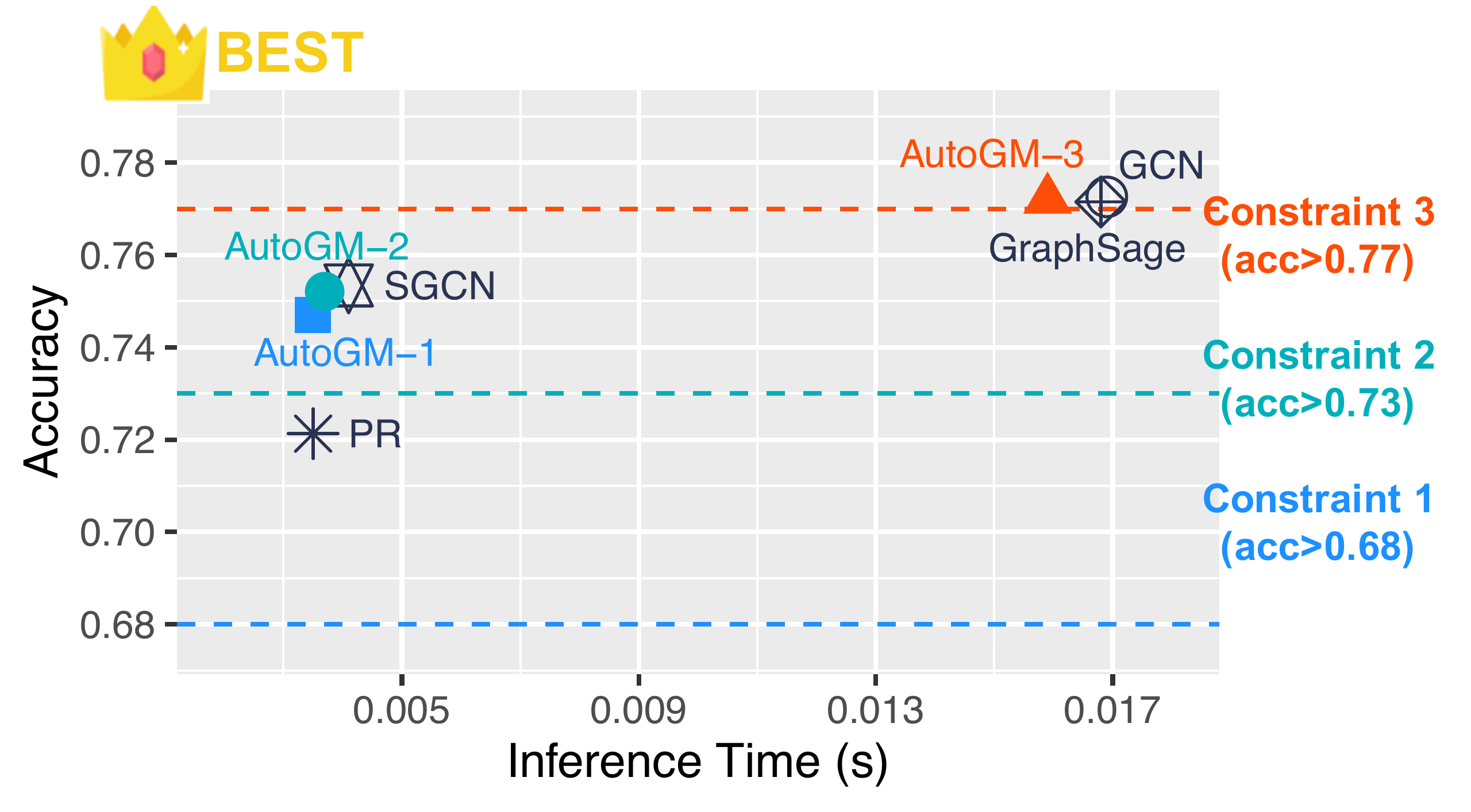}
	}
	\subfigure[Time constraints on the Pubmed dataset]
	{
		\label{fig:pubmed:time}
		\includegraphics[width=.45\linewidth]{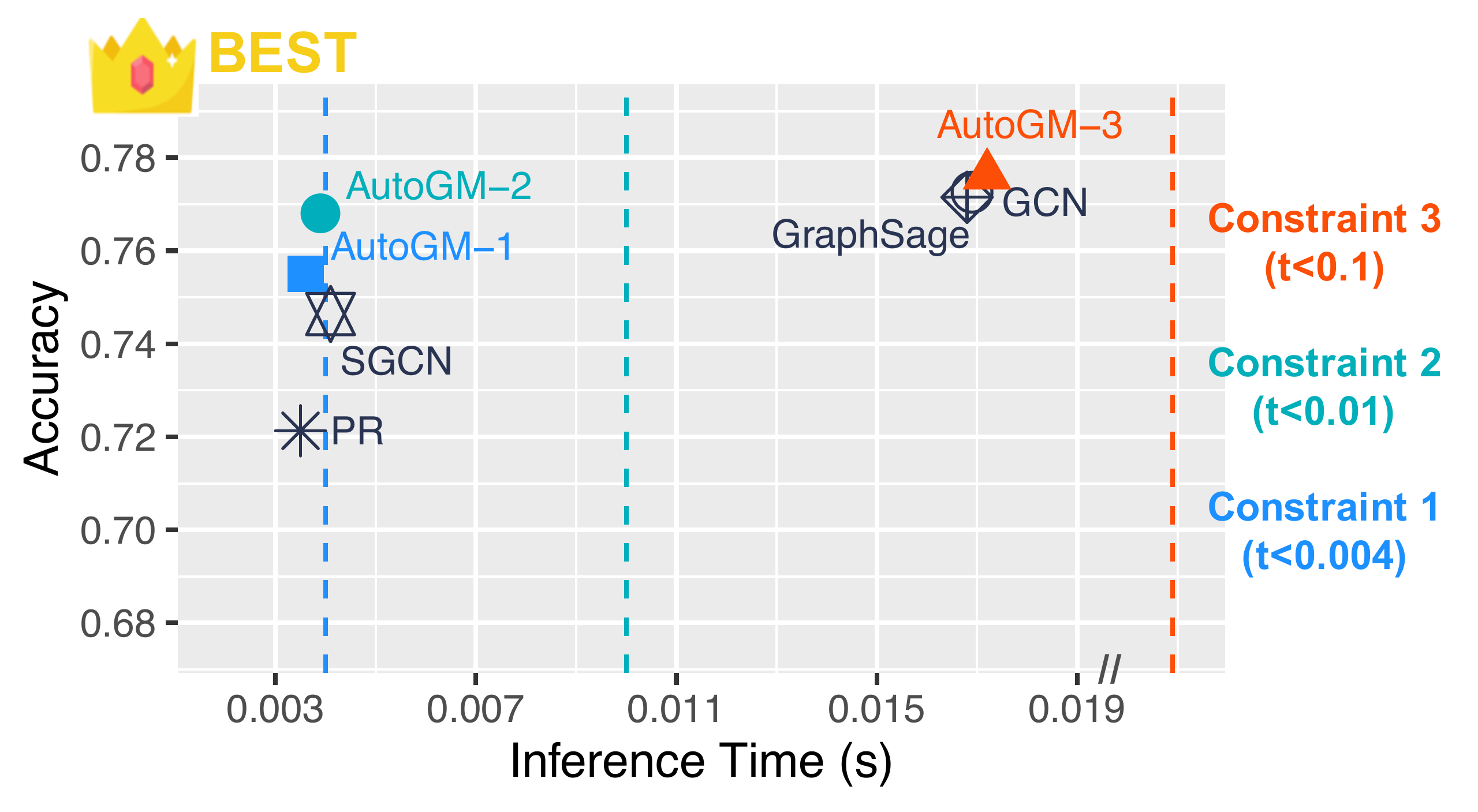}
	}
	\caption
	{ 
		\textbf{\method finds the algorithms with the best accuracy/inference time trade-off on the Cora and Pubmed datasets}:
		given three different accuracy/inference time constraints $1,2,3$, \method generates three novel graph algorithms, \method-$1,2,3$, respectively.
	}
	\label{fig:tradeoff}
\end{figure*}

In this section, we evaluate the performance of \method compared to existing models with heuristic parameters. We aim to answer the following questions:

\begin{itemize} [leftmargin=10pt]
\item {\textbf{Q1. Effectiveness of \method:}} 
Do algorithms found by \method outperform their state-of-the-art competitors? 
Given an upper bound on inference time/a lower bound on accuracy, does \method find the algorithm with the best accuracy/the fastest inference time? (Section~\ref{sec:exp:effect})
\item {\textbf{Q2. Search efficiency of \method:}} 
How long does \method take to find the optimal graph algorithm?
How efficient it is compared to random search?
 (Section~\ref{sec:exp:search})
\item {\textbf{Q3. Effect of \frame parameters:}} 
How do parameters $(d, k, w, l ,a$) affect the accuracy and inference time of a graph mining algorithm? (Section~\ref{sec:exp:parameter})
\end{itemize}

\subsection{Experimental Setting}

We evaluate the performance of graph mining algorithms on a semi-supervised node classification task. 
All experiments were conducted on identical machines using the Amazon EC2 service (p2.xlarge with 4 vCPUs, 1 GPU and 61 GB RAM).

\noindent\textbf{Dataset:} We use the three citation networks (Cora, Citeseer, and Pubmed)~\cite{sen2008collective}, two Amazon co-purchase graphs (Amazon Computers and Amazon Photo)~\cite{shchur2018pitfalls}, and two co-authorship graphs (MS CoauthorCS and MS CoauthorPhysics)~\cite{shchur2018pitfalls}.
We report their statistics in Table~\ref{tab:dataset}. 

\noindent\textbf{Baseline:} Our baselines are PageRank~\cite{page1999pagerank}, %Pixie~\cite{eksombatchai2018pixie},
GCN~\cite{kipf2016semi}, GraphSage~\cite{hamilton2017inductive}, and SGCN~\cite{wu2019simplifying}.
We generate each algorithm under \frame by setting the five parameters as follows:
\begin{itemize}[topsep=0pt, noitemsep]
\item PageRank: $d = 1, k = 30, w = -1, l = \text{False}, a = \text{NA}$
%\item Pixie: $d = 1, k = 10, w = 25, l = \text{False}, a = \text{NN}$
\item GCN: $d = 64, k = 2, w = -1, l = \text{True}, a = \text{SS}$
\item GraphSAGE: $d = 64, k = 2, w = 25, l = \text{True}, a = \text{SA}$
\item SGCN: $d = 64, k = 2, w = -1, l = \text{False}, a = \text{SS}$
\end{itemize}
\noindent 
When $w$ is larger than the number of neighbors, we sample neighbors with replacement.
For PageRank, the original algorithm outputs the sum of intermediate scores that each node receives ($\sum X_i$), but we use only the final scores $X_k$ in our experiments.
The goal of our experiments is to compare PageRank with other algorithms in terms of its main feature in \frame, low dimension ($d=1$).

\noindent\textbf{Bayesian optimization:}
We use an open-sourced Bayesian optimization package\footnote{https://github.com/fmfn/BayesianOptimization}.
For the parameters $d$, $k$, and $w$ which take integer values, we round the real-valued parameters chosen by BO to integer values. 
For the parameter $l$ and $a$, which take boolean and categorical values, we bound the search space (\small$0<l<1$\normalsize and \small$0<a<6$\normalsize), round the real-valued parameters chosen by BO to the closest integer values, and map ($0$: False, $1$: True, $0$: NN, $1$: NS, $2$: NA, $3$: SN, $4$: SS, $5$: SA).
We set the BO search budget (total number of evaluations) as $20$ for all datasets.
The resulting search time of each dataset is reported in Table~\ref{tab:efficiency}.
For the penalty coefficient $\lambda$, the smaller $\lambda$ brings the tighter budget constraints.
To make our budget constraints strict, we set $\lambda$ as \small$10^{-19}$\normalsize.

We use the Adam optimizer \cite{kingma2014adam} and tune each baseline with a grid search on each dataset. 
Most baselines perform best on most datasets with a learning rate of $0.01$, weight decay of \small$5\times10^{-4}$\normalsize, and dropout probability of $0.5$. 
We fix these parameters in our autonomous graph mining algorithm search through Bayesian Optimization. 
We report the average performance across $10$ runs for each experiment.

\begin{table}[]
	\begin{threeparttable}[]
		\centering
		\caption{Dataset statistics: AmazonC and AmazonP denote the Amazon Computer and Amazon Photo datasets, respectively.
			CoauthorC and CoauthorP denote the MS Coauthor CS and Physics, repectively.}
		\begin{tabular}{l|ccccc}
			\toprule
			\textbf{Dataset} & \textbf{Node} & \textbf{Edge} & \textbf{Feature} & \textbf{Label} &\textbf{Train/Val/Test} \\
			\midrule
			Cora&    2,485&    5,069&	1,433 & 7 & 140/500/1,000\\
			Citeseer&    2,110&    3,668& 3,703 & 6 & 120/500/1,000 \\
			Pubmed&    19,717&    44,324& 500 & 3	&60/500/1,000 \\
			AmazonC&    13,381&    245,778& 767 & 10	&410/1,380/12,000 \\
			AmazonP&    7,487&    119,043& 745 & 8	&230/760/6,650 \\
			CoauthorC&    18,333&    81,894& 6,805 & 15	&550/1,830/15,950 \\
			CoauthorP&    34,493&    247,962& 8,415 & 5	&1,030/3,450/30,010 \\
			\bottomrule
		\end{tabular}
		\label{tab:dataset}
	\end{threeparttable}
\end{table}

\subsection{Effectiveness of \method}
\label{sec:exp:effect}

In this section, we demonstrate how \method trades off accuracy and inference time in practice.
We compare the best algorithms found by \method with baselines in terms of accuracy and inference time.
For each dataset, we run \method with three different accuracy lower bounds and three inference time upper bounds, as illustrated in Figures~\ref{fig:citeseer} and~\ref{fig:tradeoff}.
For each constraint, \method generates a novel graph algorithm corresponding to a set of five parameters of \frame.
For space efficiency, we show the result on the Cora, Citeseer, and Pubmed datasets.
Performance on other datasets is reported in Table~\ref{tab:efficiency}.

Among algorithms satisfying an accuracy lower bound, the algorithms generated by \method show the best trade-off between accuracy and inference time.
For instance, in the Citeseer dataset in Figure~\ref{fig:citeseer:acc}, \method-2 has the fastest inference time above accuracy constraint $2$ among PageRank (PR), GCN, SGCN, and GraphSage. 
Given the highest or tightest accuracy constraint $3$, only \method-3 satisfies it.
Conversely, among algorithms satisfying inference time upper bounds, the algorithms generated by \method have the highest accuracy.
For instance, in the Pubmed dataset in Figure~\ref{fig:pubmed:time}, \method-1 has the highest accuracy below time constraint $1$ among PR and SGCN. 
Given the most generous time constraint $3$, \method-3 achieves the highest accuracy among all algorithms.

\begin{table}[]
	\caption{Parameters corresponding to algorithms found by \method in Figures~\ref{fig:citeseer}. The Budget column denotes the constraint input to \method to generate an algorithm.
	}
	\label{tab:tradeoff}
	\begin{tabular}{lllllll|ll}
		\toprule
		\textbf{Dataset} & \multicolumn{1}{|l|}{\textbf{Budget}} & $d$   &$k$ & $w$  & $l$& $a$ & \textbf{Time}   & \textbf{Acc} \\ \midrule
		\multirow{6}{*}{\textbf{Citeseer}}     
		& \multicolumn{1}{|l|}{t\textless{}0.004}   & 70  & 4 & 25 & F & SA & 0.0039 & 0.674  \\
		& \multicolumn{1}{|l|}{t\textless{}0.01}    & 255  & 4 & 45 & F & SS & 0.004 & 0.683  \\
		& \multicolumn{1}{|l|}{t\textless{}0.1}     & 68 & 1 & 47 & T & SS & 0.0134  & 0.686  \\ \cline{2-9} 
		& \multicolumn{1}{|l|}{a\textgreater{}0.58} & 138  & 1 & 36 & F & SA & 0.0039 & 0.622  \\
		& \multicolumn{1}{|l|}{a\textgreater{}0.63} & 25 & 4 & 54 & F & NA & 0.0039 & 0.665  \\
		& \multicolumn{1}{|l|}{a\textgreater{}0.68} & 39  & 1 & 10 & T & SS & 0.0121  & 0.69   \\
		\bottomrule
	\end{tabular}
\end{table}

\begin{table*}[]
	\vspace{-5mm}
	\caption{
	Search efficiency of \method: given the same search time (column 2) and accuracy lower bounds (column 3), \method finds faster algorithms than RandomSearch across all datasets; similarly, given the same search time (column 2) and inference time upper bounds (column 8), \method finds more accurate algorithms than RandomSearch across all datasets.
	}
	\label{tab:efficiency}
	\begin{tabular}{l|c|c|cccc|c|cccc}
		\toprule
		&                        &                   & \multicolumn{2}{c}{\textbf{Fastest Inference (s)}} & \multicolumn{2}{c|}{\textbf{Accuracy}} &          & \multicolumn{2}{c}{\textbf{Highest Accuracy}} & \multicolumn{2}{c}{\textbf{Inference (s)}} \\
		\textbf{Dataset}            & \textbf{Search(s)} & \textbf{Min.Acc.} & \textbf{AutoGM}      & \textbf{Random}      & \textbf{AutoGM}   & \textbf{Random}  & \textbf{Max.Time(s)} & \textbf{AutoGM}     & \textbf{Random}    & \textbf{AutoGM}     & \textbf{Random}    \\ \midrule
		\textbf{Cora}      & 450                    & 0.78              & \textbf{0.0034}               & -                    & 0.79             & -                & 0.004             & \textbf{0.77}                & \textbf{0.77}               & 0.0036              & 0.0033             \\
		\textbf{Citeseer}  & 800                    & 0.67              & \textbf{0.0039}               & \textbf{0.0039}               & 0.67              & 0.67             & 0.004             & \textbf{0.67}                & -                  & 0.0039              & -                  \\
		\textbf{Pubmed}    & 1,800                  & 0.75              & \textbf{0.021}                & -                    & 0.77              & -                & 0.004             & \textbf{0.76}                & 0.71               & 0.0036              & 0.0039             \\
		\textbf{AmazonC}   & 5,700                  & 0.85              & \textbf{0.032}                & 0.033                & 0.89              & 0.87             & 0.04              & \textbf{0.85}                & -                  & 0.032               & -                  \\
		\textbf{AmazonP}   & 18,000                 & 0.93              & \textbf{0.047}                & 0.065                & 0.94              & 0.93            & 0.05              & \textbf{0.94}                & -                  & 0.048               & -                  \\
		\textbf{CoauthorC} & 2,500                  & 0.8               & \textbf{0.015}                & 0.016                & 0.8               & 0.82             & 0.02              & \textbf{0.83}                & 0.75               & 0.015               & 0.02               \\
		\textbf{CoauthorP} & 1,500                  & 0.9               & \textbf{0.01}                 & -                    & 0.91              & -                & 0.01              & \textbf{0.92}                & 0.86               & 0.01                & 0.01              \\\bottomrule
	\end{tabular}
\end{table*}

The empirical performance of our baselines is consistent with our guidelines for how to choose the parameters ($d, k, w, l, a$) in Section \ref{sec:frame:parameter}.
PageRank achieves fast inference time with a low dimension of messages ($d=1$) and no nonlinearities ($l=\text{False}$), but sacrifice accuracy. 
%Pixie underperforms because it is the only algorithm where messages are aggregated without normalization. 
%This aggregation strategy is not well-suited to our semi-supervised node classification task because it leads to node embeddings monotonically increasing with every aggregation step.
GCN and GraphSage achieve high accuracy with a high dimension of messages ($d=64$) and nonlinearities ($l=\text{True}$) at the cost of a high inference time.
SGCN removes nonlinearities ($l=\text{False}$) to decrease the inference time while maintaining high accuracy.

Table~\ref{tab:tradeoff} shows the parameter set of \frame that corresponds to the algorithms found by \method on the Citeseer dataset.
When encouraged to find higher accuracy algorithms (through a larger time upper bound or higher accuracy lower bound), \method is likely to use high values of $d$ and $w$ and nonlinearities ($l=\text{True}$).
For instance, \method chooses higher values $d = 255, w=45$ for the larger time upper bound  $time < 0.01$ than the values $d = 70, w=25$ for the bound $time < 0.004$.
With the largest upper bound $time < 0.1$, \method chooses $l=True$ to use nonlinearities. 
This result is consistent with our intuition over the parameter selection in Section~\ref{sec:frame:parameter}.
Vastly different parameter sets for each algorithm in Table~\ref{tab:tradeoff} show that \method searches the parameter space beyond human intuition, which underlines the value of autonomous graph mining algorithm development.

\subsection{Search efficiency of \method}
\label{sec:exp:search}

\method searches for the optimal graph algorithm in a five-dimensional space $(d, k, w, l, a)$ defined by \frame.
To show the search efficiency of \method, we give the same maximum search time and budget constraints to \method and RandomSearch, then compare the performance of the best graph algorithms each method finds.
RandomSearch samples each parameter $(d, k, w, l, a)$ randomly and defines a graph algorithm based on the sampled parameters.
We set the maximum search time proportional to the size of the dataset.
The budget constraints are chosen based on the best performance among the baseline methods (PageRank, GCN, GraphSage, SGCN).
We select the tightest constraints (i.e., fastest inference time and highest accuracy among the baselines) to examine the search efficiency.

Table~\ref{tab:efficiency} shows the inference time and accuracy of the optimal graph algorithms \method and RandomSearch find.
RandomSearch fails to find any algorithm satisfying the given accuracy constraints on the Cora, Pubmed, and CoauthorP datasets.
It also fails to find any algorithm satisfying the inference time constraints on the Citeseer, AmazonC, and AmazonP datasets.
When RandomSearch finds graph algorithms satisfying the given constraints, their performance is still lower than the algorithms found by \method.
For instance, given the inference time upper bound ($t<0.02$) on the CoauthorC dataset, \method finds an algorithm with accuracy $0.83$ while RandomSearch finds an algorithm with accuracy $0.75$.

Table~\ref{tab:efficiency} presents how much accuracy/inference time is used under the given budgets to find the optimal graph algorithms (column $6,7$ and $11,12$).
\method generates algorithms whose accuracy (time) is as close as possible to the given accuracy (time) budgets.
For instance, \method finds the fastest graph algorithm with an accuracy of $0.8$ when the accuracy lower bound is given as $0.8$ on the CoauthorC dataset.
By exhausting the budget, \method improves the target metric time (accuracy) and brings the best trade-off between computation time and accuracy.

\vspace{-1mm}
\subsection{Effect of \frame parameters}
\label{sec:exp:parameter}

\begin{figure}[t!]
	\centering
	\includegraphics[width=.6\linewidth]{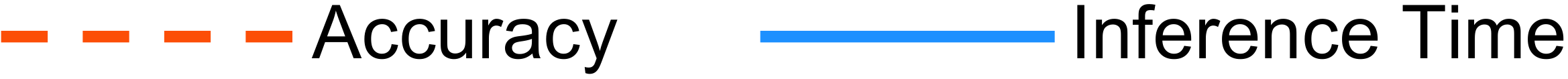}
	\subfigure[Parameter $d$]
	{
		\label{fig:param_d}
		\includegraphics[width=.43\linewidth]{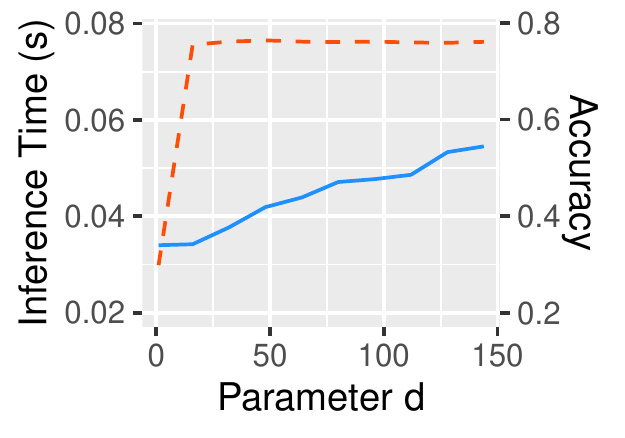}
	}
	\subfigure[Parameter $k$]
	{
		\label{fig:param_k}
		\includegraphics[width=.43\linewidth]{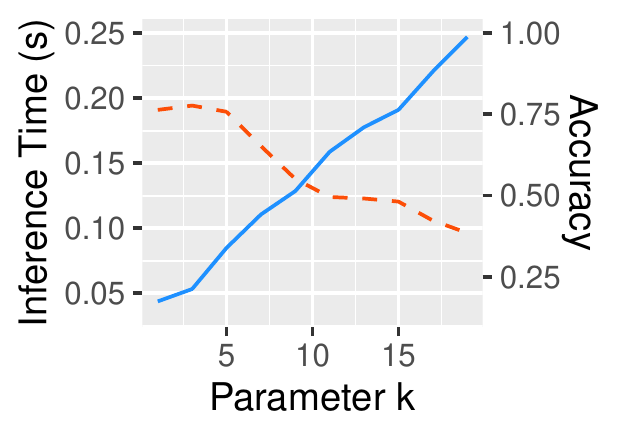}
	}
	\subfigure[Parameter $w$]
	{
		\label{fig:param_w}
		\includegraphics[width=.43\linewidth]{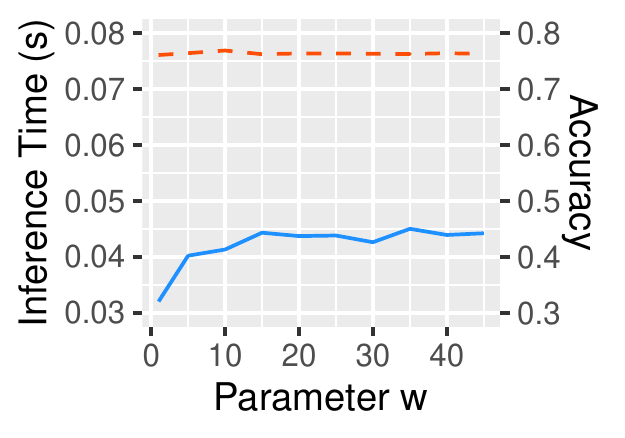}
	}
	\subfigure[Parameter $l$]
	{
		\label{fig:param_l}
		\includegraphics[width=.43\linewidth]{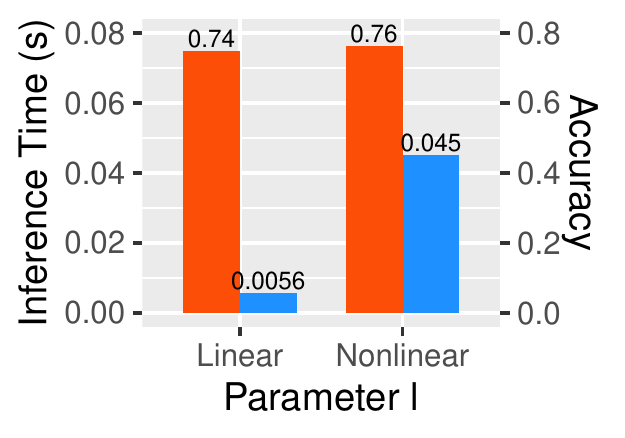}
	}
	\includegraphics[width=.6\linewidth]{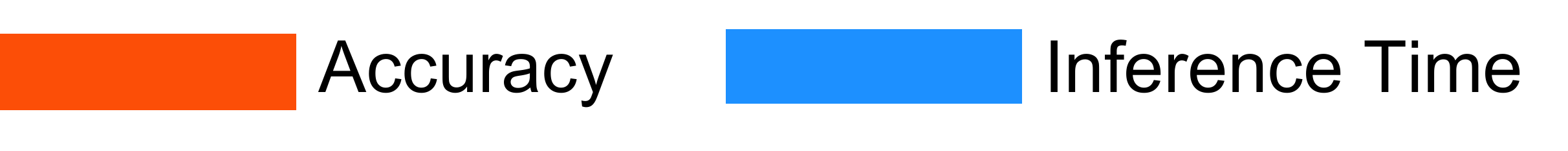}
	\subfigure[Parameter $a$]
	{
		\label{fig:param_a}
		\includegraphics[width=.8\linewidth]{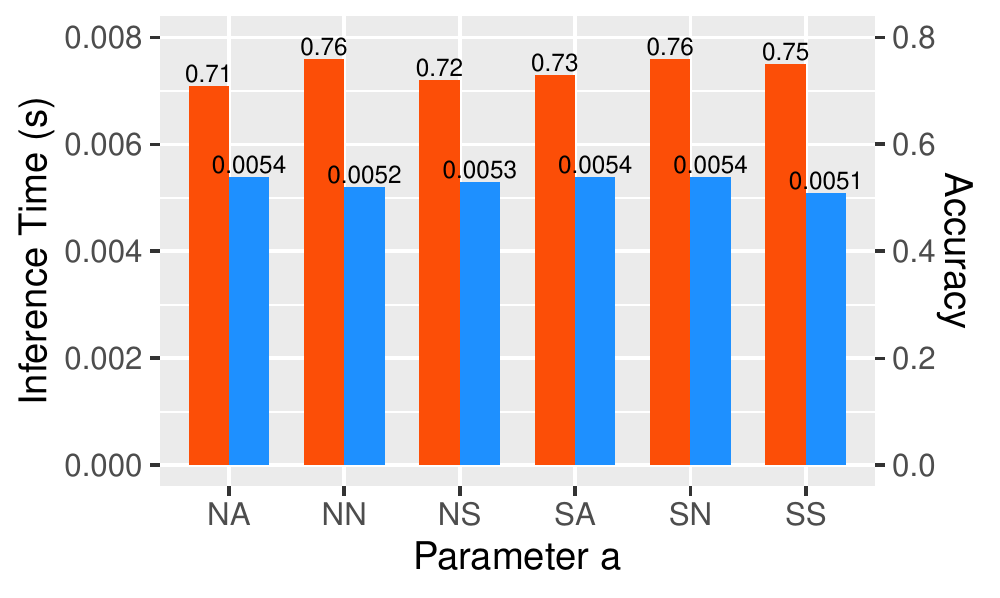}
	}
	\caption
	{ 
		Effects of the five parameters ($d, k, w, l, a$) of \frame on the performance of graph algorithms (i.e. accuracy and time).
	}
	\label{fig:param}
\end{figure}

In this section, we investigate the effects of parameters of \frame on the performance of a graph mining algorithm.
Given a set of parameters (\small$d = 64, k = 2, w = -1, l = \text{True}, a = \text{SS}$\normalsize), we vary one parameter  while fixing the others and measure the performance of the generated algorithm. 
For the experiment where we vary the aggregation parameter $a$, we use a different set of parameters (\small$d =16, k = 2, w = 10, l = False$\normalsize) to better illustrate changes in accuracy and inference time.
For brevity, we show the result on the Pubmed dataset.

\begin{itemize}[leftmargin=10pt]
	\item {
		\textbf{Dimension $d$:} 
		Figure~\ref{fig:param_d} shows that inference time increases linearly with $d$, while accuracy increases only until $d > 20$.
		For the Pubmed dataset, $20$-dimensional messages are expressive enough that the accuracy stops increasing.
		Larger datasets would likely benefit from higher dimensional messages. 
	}
	\item {
		\textbf{Length $k$:}
		In Figure~\ref{fig:param_k}, when $k$ increases, inference time increases linearly, but accuracy decreases for $k > 3$.
		The decrease in accuracy is due to oversmoothing: repeated graph aggregations eventually make node embeddings indistinguishable.
	}
	\item {
		\textbf{Width $w$:} 
		In Figure~\ref{fig:param_w}, when $w$ increases, inference time increases until $w > 15$, but accuracy does not change noticeably.
		The plateau in accuracy is due to most nodes having few neighbors and nearby nodes sharing similar feature information, which makes a single sampled node be a representative of a node's whole neighborhood.
		The plateau in inference time indicates that nodes have fewer than $15$ neighbors on average on the Pubmed dataset.
	}
	\item {
		\textbf{Nonlinearity $l$:}
		Figure~\ref{fig:param_l} shows that adding nonlinearities ($l = \text{True}$) increases accuracy due to richer expressiveness, but also inference time.
	}
	\item {
	\textbf{Aggregation strategy $a$:}
		Figure~\ref{fig:param_a} shows that the choice of aggregation strategy $a$ has a considerable effect on the accuracy of a graph mining algorithm. Still, we cannot conclude that any aggregation strategy is always superior to others.
	}
\end{itemize}

\noindent
Figure~\ref{fig:param} shows the general tendency in the effects of the parameters.
Different datasets have slightly different results (e.g., which $w$ stops increasing accuracy or which $k$ starts bringing oversmoothing).
This shows the need for \method, which chooses the best parameter set automatically for the dataset we employ. 

\section{Conclusion \& Future Work}
\label{sec:conclusion}
In this paper, we introduce an automated system \method for graph mining algorithm development. 
Our main contributions are:
\begin{itemize}[leftmargin=10pt]
	\item {
		\textbf{Unification:}
		\frame allows conventional and GNN algorithms to be unified in the same framework for the first time, which is necessary to establish the parameter space for algorithm search.
	}
	\item {
		\textbf{Automation:}
		Based on the search space defined by \frame, \method finds the optimal graph algorithm using Bayesian optimization.
	}
	\item {
		\textbf{Budget awareness:}
		\method maximizes the performance of an algorithm under a given time/accuracy budget.
	}
	\item {
		\textbf{Effectiveness:}
		\method finds novel graph algorithms with the best speed/accuracy trade-off on real-world datasets.
	}
\end{itemize}

\noindent We hope this paper will spark further research in this direction and empower practitioners without much expertise in graph mining to deploy graph algorithms tailored to their scenarios. 
%Future works include extending \method to new kinds of algorithms (e.g., message passing based tensor algorithms, algorithms for time-evolving graphs).

\bibliography{BIB/myref}
\bibliographystyle{abbrv}

\end{document}